\documentclass[final,3p,times]{elsarticle}

\makeatletter
\def\ps@pprintTitle{
  \let\@oddhead\@empty
  \let\@evenhead\@empty
  \let\@oddfoot\@empty
  \let\@evenfoot\@empty}
\makeatother

\usepackage{amssymb}
\usepackage{amsmath}
\usepackage{hyperref}

\begin{document}

\begin{frontmatter}

\title{FIGNN: Feature-Specific Interpretability for Graph Neural Network Surrogate Models}

\author[1]{Riddhiman Raut\corref{cor1}}
\author[2]{Romit Maulik}
\author[3]{Shivam Barwey}

\address[1]{Department of Mechanical Engineering, Pennsylvania State University, University Park, 16802, PA, USA}
\address[2]{College of Information Sciences and Technology, Pennsylvania State University, University Park, 16802, PA, USA}
\address[3]{Transportation and Power Systems Division, Argonne National Laboratory, Lemont, 60439, IL, USA}

\begin{abstract}
This work presents a novel graph neural network (GNN) architecture, the Feature-specific Interpretable Graph Neural Network (FIGNN), designed to enhance the interpretability of deep learning surrogate models defined on unstructured grids in scientific applications. Traditional GNNs often obscure the distinct spatial influences of different features in multivariate prediction tasks. FIGNN addresses this limitation by introducing a feature-specific pooling strategy, which enables independent attribution of spatial importance for each predicted variable. Additionally, a mask-based regularization term is incorporated into the training objective to explicitly encourage alignment between interpretability and predictive error, promoting localized attribution of model performance. The method is evaluated for surrogate modeling of two physically distinct systems: the SPEEDY atmospheric circulation model and the backward-facing step (BFS) fluid dynamics benchmark. Results demonstrate that FIGNN achieves competitive predictive performance while revealing physically meaningful spatial patterns unique to each feature. Analysis of rollout stability, feature-wise error budgets, and spatial mask overlays confirm the utility of FIGNN as a general-purpose framework for interpretable surrogate modeling in complex physical domains.
\end{abstract}


\end{frontmatter}

\section{Introduction}

Graph Neural Networks (GNNs) have become increasingly prominent in modeling systems governed by partial differential equations (PDEs)~\cite{li2020multipole, wu2022graph, bronstein2017geometric, kurz2025harnessing, horie2022physics, han2022predicting}. By representing discretized spatial domains—such as structured grids or unstructured meshes—as graphs, GNNs (through learned message passing operations \cite{gilmer}) can be trained to model complex non-local interactions conditioned on node neighborhood connectivities contained in the input graph. This paradigm allows GNNs to generalize across varying mesh resolutions and geometries~\cite{khan2024graphmesh}, making them especially advantageous in PDE-based surrogate and closure modeling applications \cite{romit_review}.

The flexibility of GNN-based surrogate modeling frameworks has driven their adoption across various scientific and engineering domains, particularly in scenarios where models must not only capture complex dynamics/physics but also be compatible with unstructured data formats. Foundational GNN architectures, including (but not limited to) Graph Convolutional Networks~\cite{kipf2016semi}, GraphSAGE ~\cite{hamilton2017inductive}, graph attention networks \cite{gat}, and the broader framework of mesh-based simulations using GNNs~\cite{battaglia2018relational, sanchez2020learning, pfaff2020learning}, have demonstrated impressive capabilities in capturing complex relational data. A key example for spatiotemporal modeling is MeshGraphNet~\cite{pfaff2020learning}, which leverages message passing GNNs to model fluid dynamics phenomena on unstructured meshes. Similar approaches, including multiscale extensions (characterized by employing message passing operations at multiple length scales within the model architecture) have since been developed to construct improved models for not only unsteady mesh-based fluid flows \cite{fortunato2022multiscale,lino2022multi,shivam_gnn_jcp}, but also structural mechanics \cite{deshpande2022magnet,perera2024multiscale}, weather forecasting \cite{lam2023learning}, molecular dynamics \cite{li_gnn_md,park2024scalable}, and fluid-structure interactions \cite{jaiman_fsi_gnn}. Additionally, identification of parallels between the action of GNN layers on meshes and physics-based PDE discretization procedures have been used to construct numerics-informed GNN models aligned with finite volume \cite{karthik_neurips_gnn,li2024predicting} and finite/spectral element \cite{jianxun_gnn_galerkin,jaiman_phignn,sb_srgnn,physgnn} formulations. 

Despite their advantages, GNN surrogates largely remain black-box models, posing a significant hurdle to their widespread adoption in engineering applications. While effective in modeling complex physical systems, these models offer limited transparency into their internal reasoning, leaving users unable to clearly interpret what the networks have learned, why particular spatial regions or features influence predictions, and how latent representations correspond to physical phenomena~\cite{schmidt2024towards}. Post‑hoc explainability tools -- such as Grad‑CAM for images~\cite{selvaraju2020grad}, GNNExplainer~\cite{ying2019gnnexplainer}, PGExplainer~\cite{luo2020parameterized}, GraphLIME~\cite{huang2022graphlime}, or saliency maps for graph data \cite{harel2006graph} -- can be used to add interpretability onto an already‑trained network in a postprocessing stage. These methods typically provide activation heatmap visualizations \cite{selvaraju2020grad,harel2006graph} and, in GNN-specific contexts, post-hoc extraction of important sub-graphs and/or node feature subsets \cite{ying2019gnnexplainer,luo2020parameterized,huang2022graphlime}. Although useful for qualitative analysis, such approaches introduce a key drawback in that the interpretability workflow is decoupled from the optimization process used to construct the model, which can skew/alter physical analysis and retains the black-box nature of models. Furthermore, the analyses obtained through such techniques also lacks a quantitative connection with the a-posteriori performance of the trained surrogate model and requires subjective interpretation from the end-user.

An alternative approach is to overcome this limitation by baking in interpretability within the model itself (referred to as active interpretability \cite{zhang2021survey}). Embedding interpretability inside the forward pass of a surrogate offers three decisive advantages. First, the explanatory mechanism is co‑optimised with the task loss, guaranteeing that whatever the model marks as ``important'' is influential for prediction, rather than a post‑hoc artifact. Second, explanations are produced directly at inference time: a single forward call simultaneously yields the forecast and a compact representation (mask, sub‑graph, attention field) of the regions driving that forecast, eliminating costly per‑sample optimization loops. Third, if the explanatory module is differentiable, one can add auxiliary objectives -- e.g. sparsity, error‑budget constraints, or physics‑aware regularization -- to steer the model toward features that are both concise and physically meaningful, enabling downstream actions such as adaptive meshing or sensor deployment. For GNN surrogate modeling of fluid dynamics, Barwey et al.~\cite{barwey2023multiscale} demonstrated the efficacy of this method by appending a differentiable Top‑K pooling layer~\cite{gao2019graph} to a multiscale GNN; the layer adaptively subsamples nodes during training, producing a global spatial mask that highlights the sub‑graphs most responsible for forecasting error in unsteady fluid‑flow data. Because the mask is learned jointly with the surrogate, it doubles as an a‑posteriori error indicator: using a budget regularization strategy, graph nodes identified by the pooling operation can identify where the GNN is expected to incur the largest forecasting residual, allowing subsequent targeted refinement \cite{barwey2025interpretable}. These studies highlight how interpretable GNNs can expose physically coherent structures in fluid dynamics -- recirculation zones in a backward‑facing‑step (BFS) flow, for example -- and can even signal where numerical error will accumulate during inference. 


However, the above studies share one crucial limitation: they yield a single global mask/feature tuned to minimize the aggregate loss over all node features (i.e., the interpretability component of the model is entangled with all input node features). A single global mask inevitably blurs the distinct spatial signatures that drive each variable, hampering physical insight and any downstream task that demands feature-wise attribution -- adaptive sensor placement for humidity versus temperature, error control for pressure versus velocity, and so on -- which are critical to distinguish in multi-physics applications. This calls for a feature‑specific, jointly‑trained masking framework that preserves the benefits of built‑in interpretability while disentangling the contributions of disparate physical quantities. This article aims to address this research gap by introducing Feature‑specific Interpretable Graph Neural Networks (FIGNN). Here, the surrogate architecture learns a dedicated Top‑K mask for every node variable, enabling variable‑by‑variable analysis, error‑tagging and domain adaptation without compromising predictive accuracy. They key contributions of this work are as follows:

\begin{itemize}
\item A feature-specific interpretability module (FSIM) is introduced - that spawns an independent, differentiable mask for every node feature—disentangling the spatial drivers of heterogeneous physics.  The block interfaces with any message‑passing baseline (demonstrated on a multiscale GNN in this study), making the framework architecture‑agnostic and readily reusable across domains.
\item During training the baseline surrogate is frozen and only the FSIM branch(es) are optimized. At inference, a single forward pass simultaneously delivers (i) the multi‑variable forecast and (ii) a stack of feature‑specific masks.This keeps runtime identical to the frozen baseline and avoids the heavy per‑sample cost typical of post‑hoc explainers.
\item The framework is tested on two contrasting datasets—one representing the short-term evolution of the atmosphere obtained via the SPEEDY emulator \cite{kucharski2006decadal} (structured grid, multi‑physics) and unstructured OpenFOAM backward‑facing‑step (BFS) flow. FIGNN consistently isolates coherent, variable‑specific structures, demonstrating robustness to mesh type, scale separation, and governing equations.
\item The error‑budget regularization technique \cite{barwey2025interpretable} is extended to the multi‑mask setting, showing that a lightweight budget term guides each feature‑mask to contain and localize its own variable-specific forecast error. This dual role—interpretation + error tagging—enables downstream tasks such as adaptive mesh refinement or sensor placement to be driven by feature‑wise uncertainty cues.

\end{itemize}

The remainder of this manuscript is structured as follows: Section 2 details the proposed methodology—introducing the FIGNN architecture, training procedure, and the two benchmark datasets (SPEEDY climate and BFS flow). Section 3 presents a thorough evaluation, including accuracy, mask interpretability, and error‑tagging results on both datasets. Section 4 concludes with the main take‑aways and outlines promising avenues for future work.

    
    
    
    

\section{Methodology}

The forecasting problem is posed on a static spatio-temporal graph $G = (V, E)$, leveraging graph neural networks (GNNs) for fluid dynamics forecasting as demonstrated in recent literature~\cite{gao2024predicting, han2022predicting}. Each node $i \in V$ corresponds to a spatial sample point, while the edge set $E$ encodes geometric neighborhoods.

For each ordered edge $(i,j) \in E$, we define the invariant edge-feature vector:
\[
  \mathbf{e}_{ij} = \left[ d^{\mathrm{euc}}_{ij},\ dx_{ij},\ dy_{ij} \right], \qquad
  d^{\mathrm{euc}}_{ij} = \lVert \mathbf{p}_j - \mathbf{p}_i \rVert_2, \qquad
  (dx_{ij}, dy_{ij}) = \mathbf{p}_j - \mathbf{p}_i,
\],
where $\mathbf{p}_i$ denotes the $(x,y)$ coordinates of node $i$.

Let $X_t = [x_{1,t}, x_{2,t}, \dots, x_{N,t}]^{\top} \in \mathbb{R}^{N \times F}$ be the node-feature matrix at time $t$, with $x_{i,t} \in \mathbb{R}^F$ representing node $i$'s feature vector. The GNN $f_{\theta}$ is trained to predict feature increments through the residual update:
\begin{equation}
  X_{t+\Delta t} = X_t + f_{\theta}\!\left( X_t, G \right),
  \label{eq:predictive_task}
\end{equation}
where $\theta$ denotes learnable parameters. Both datasets employ fixed spatial relationships and edge attributes throughout training and inference phases, enabling direct interpretation of learned physical interactions without post-hoc analysis.

FIGNN is designed as an \emph{interpretability add-on}: it can be attached to any message-passing GNN that follows the widely used encode–process–decode workflow~\citep{velivckovic2019neural}. In practice, the underlying surrogate may be a standard architecture such as a Graph Convolutional Network~\citep{kipf2016semi} or GraphSAGE~\citep{hamilton2017inductive}, a physics-informed operator like PDE-GCN~\citep{eliasof2021pde}, multiscale models such as MeshGraphNets~\citep{pfaff2020learning, fortunato2022multiscale}, or even an attention-based graph transformer~\citep{yun2019graph}.  We choose the multiscale message-passing (MMP) model of ~\citep{barwey2023multiscale} in our experiments purely for performance parity with prior fluid-dynamics studies; but it is important to emphasize that FIGNN is agnostic to this choice. The baseline follows an encoder–processor–decoder layout: 2-layer multi-layer perceptrons (MLPs) handle encoding and decoding, and a single MMP block serves as the processor, performing two radius-based coarsen–refine cycles that double the characteristic edge length at each level. Within each cycle, information is aggregated using 2 message-passing layers, which are, once again, 2-layer MLPs. After the baseline model is trained, its weights are frozen; during the interpretability phase, only the parameters of the feature-specific module are updated. 

\begin{figure}[t]
    \centering
    \includegraphics[width=0.95\textwidth]{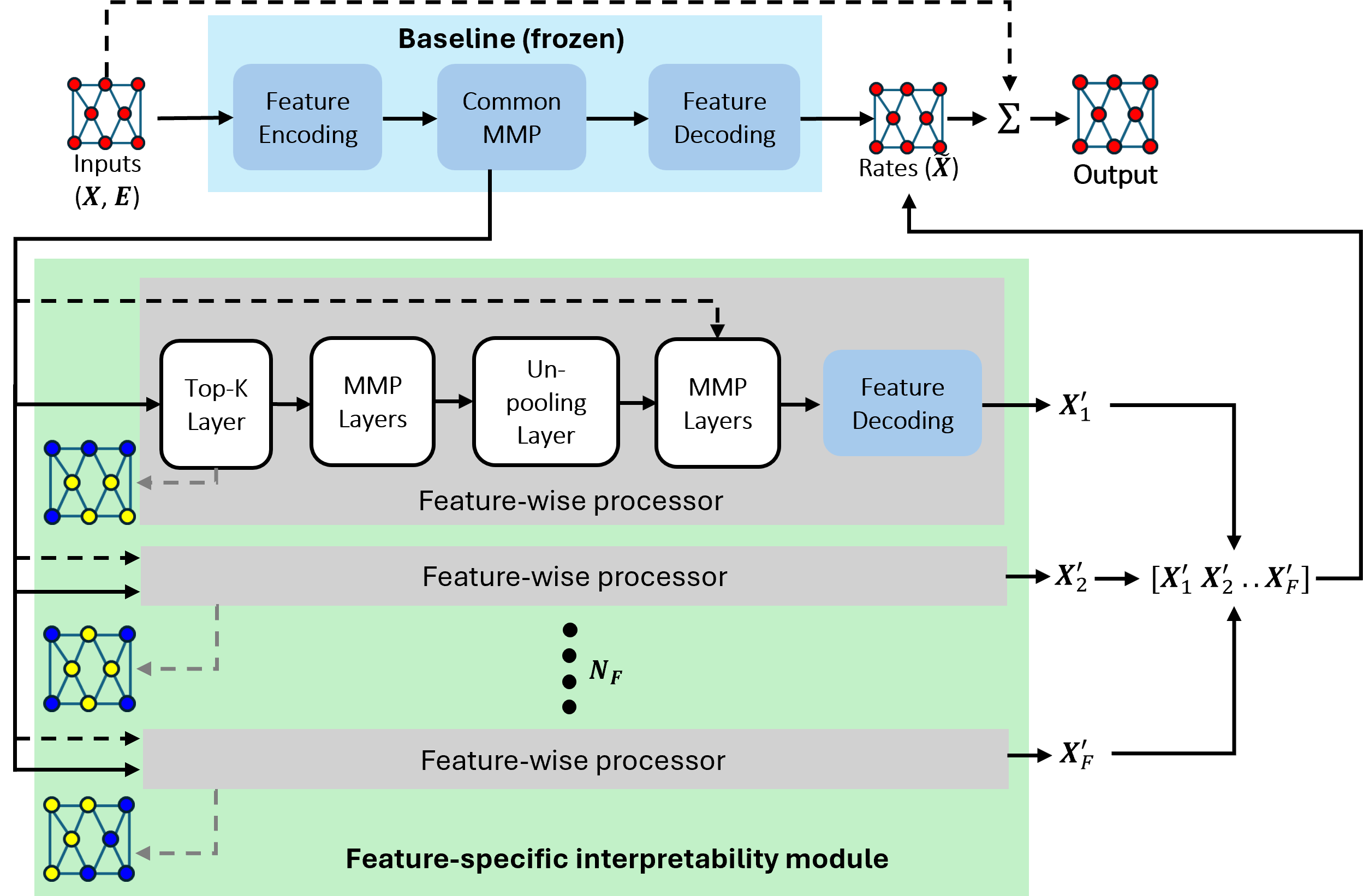}
    \caption{Schematic of the Feature-specific Interpretable Graph Neural Network (FIGNN). The frozen baseline model (blue) consists of encoding, multiscale message passing (MMP), and decoding stages. The feature-specific interpretability module (green) contains \( N_F \) parallel feature-wise processors, each with a Top-K pooling layer, MMP layers, an unpooling operation, and a frozen decoder. This design supports feature-level attribution of spatial influence.}
    \label{fig:fig_schematic}
\end{figure}

On top of this fixed surrogate we attach FIGNN’s interpretability module. Figure~\ref{fig:fig_schematic} illustrates the full FIGNN architecture. The top block represents the frozen baseline GNN model, while the lower block shows the parallel feature-wise processors in the interpretability module. This module is composed of \( N_F \) parallel \emph{feature-wise processors}, one for each output variable. These processors operate on shared latent embeddings and independently isolate spatial regions, termed masked fields, most critical to their respective features \textit{during the forward pass}. This design ensures that FIGNN provides spatial attributions without altering the predictive core, thereby preserving the integrity of the original surrogate model.

\begin{figure}
  \centering
  \includegraphics[width=\textwidth]{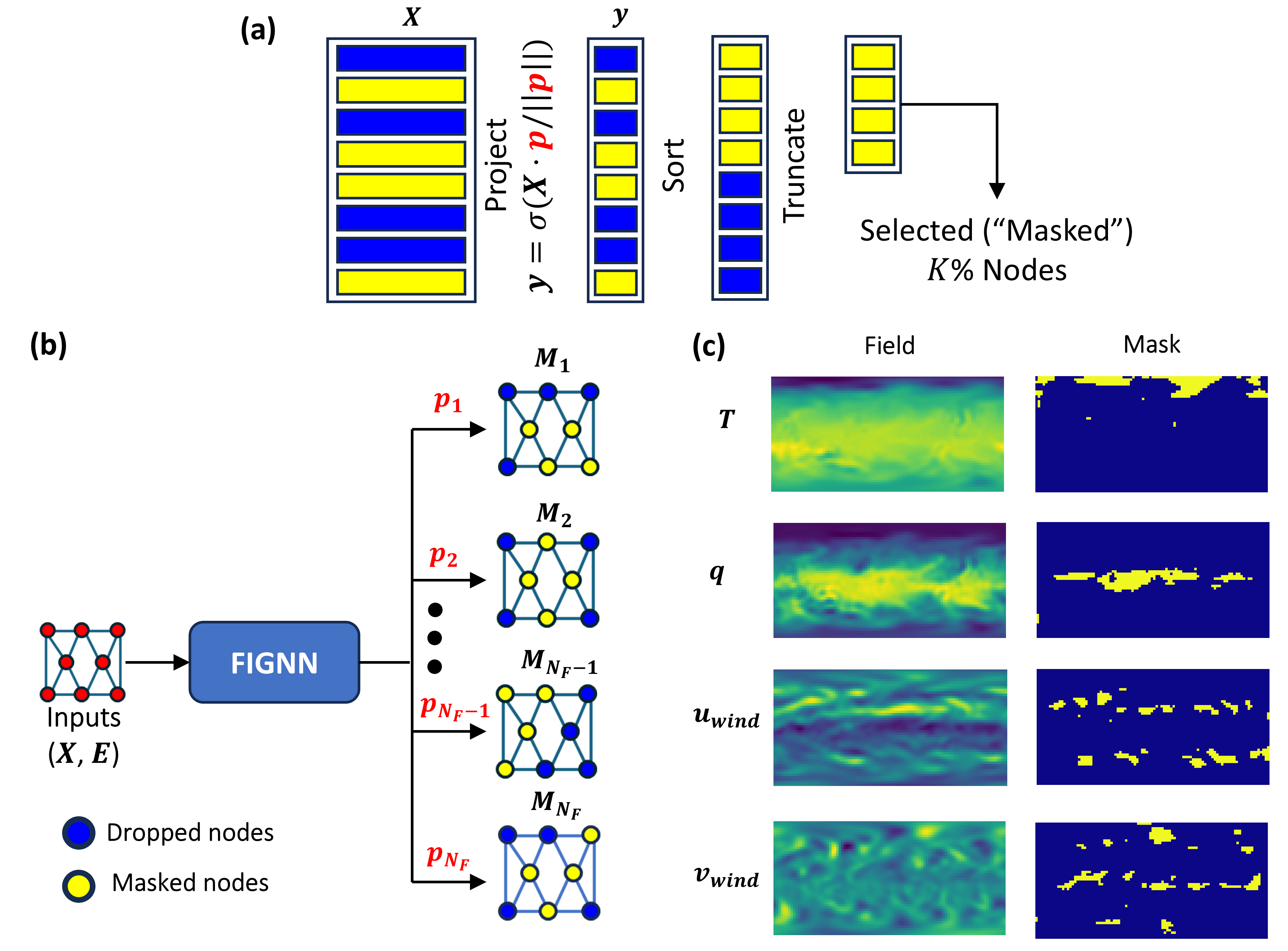}   
  \caption[\textbf{Feature‑specific Top–K pooling.}]
  {\textbf{Feature‑specific Top–K pooling and mask generation.}
  (\textbf{a})  One‑dimensional projection of the node–feature matrix
  $\mathbf X\in\mathbb{R}^{|V|\times N_F}$ onto a learnable vector
  $\mathbf p$ produces scores $\mathbf y=\sigma(\mathbf X\cdot\mathbf
  p/\lVert\mathbf p\rVert)$.  Scores are sorted and truncated, keeping
  only the highest‑ranking fraction, which defines the set of selected
  (\emph{masked}) nodes (yellow).  %
  (\textbf{b})  In FIGNN, each input feature $f\in\{1,\dots,N_F\}$
  has its own projection vector $\mathbf p_f$,  resulting in
  feature‑specific masks
  ($M_1,\dots,M_{N_F}$).
  (\textbf{c})  Example masks generated from the SPEEDY dataset (described in Sec.~\ref{sec:datasets})
  for temperature $T$, specific humidity $q$, zonal wind
  $u_{\text{wind}}$ and meridional wind $v_{\text{wind}}$; yellow =
  retained nodes, blue = dropped nodes.}
  \label{fig:topk_feature_specific}
\end{figure}

Each processor begins with a Top-K pooling layer~\citep{gao2019graph} that assigns an importance score \( s_{if} \) to each node \( i \) for feature \( f \), computed as

\begin{equation}
    s_{if} = \sigma(w_f^\top h_i),
    \label{eq:mask_score}
\end{equation}

where \(h_i \in \mathbb{R}^{128}\) is the is the node embedding produced by the frozen
processor (common MMP block) of the baseline surrogate, , \(w_f \in \mathbb{R}^{128}\) is a learnable vector for feature \( f \), and \( \sigma \) denotes the sigmoid activation. Retaining the indices of the largest \(K\%\) scores,
\(J_f=\operatorname{TopK}\bigl(s_{\cdot f},K\bigr)\),
defines the feature–specific sub-graph
\(G_f=(J_f,E_f)\) and the pooled representation
\(\mathbf{h}^{\mathrm{pool}}_{f}=\{h_i\;|\; i\in J_f\}\). A schematic for this pooling process is shown in Fig.~\ref{fig:topk_feature_specific}(a).
The pooled graph is then passed through an \(L_{\text{down}}\)-layer
multiscale message–passing stack,
\begin{equation}
\mathbf{h}^{\mathrm{coarse}}_{f}
\;=\;
\mathcal{M}^{\mathrm{down}}_{f}\!\bigl(
\mathbf{h}^{\mathrm{pool}}_{f},E_f\bigr).
\label{eq:mmp_down}
\end{equation}
The coarse representation is scattered back to the original node set
and added to the latent features that bypassed pooling:
\begin{equation}
\tilde{\mathbf{h}}_{f}
\;=\;
U_f\!\bigl(\mathbf{h}^{\mathrm{coarse}}_{f},J_f\bigr)
\;+\;\mathbf{h},
\label{eq:skip_add}
\end{equation}
where \(U_f\) writes each vector in
\(\mathbf{h}^{\mathrm{coarse}}_{f}\) to its parent index in \(J_f\) and
inserts zeros elsewhere.
The sum in~\eqref{eq:skip_add} is formed \emph{before} the upward pass,
so every subsequent up-layer operates on the skip-enhanced field.
A symmetric \(L_{\text{up}}\)-layer stack refines the combined
representation on the full graph:
\begin{equation}
\mathbf{h}^{\mathrm{up}}_{f}
\;=\;
\mathcal{M}^{\mathrm{up}}_{f}\!\bigl(
\tilde{\mathbf{h}}_{f},E\bigr).
\label{eq:mmp_up}
\end{equation}
Finally, the frozen baseline decoder—an MLP
\(D:\mathbb{R}^{128}\!\rightarrow\!\mathbb{R}^{N_F}\)—maps the refined
embeddings to physical space.  The contribution of processor \(f\) is
extracted from its designated channel:
\begin{equation}
x'_{i,f}
\;=\;
\bigl[D\!\bigl(\mathbf{h}^{\mathrm{up}}_{f}\bigr)\bigr]_{i,f},
\label{eq:decoder}
\end{equation}
Each branch therefore only modifies a single output feature and the full
network prediction is assembled by concatenating the
feature-wise outputs. This setup enables simultaneous, independent extraction of which nodes (and as a by-product, which regions in physical space) contribute most to each predicted variable, which in turn facilitates feature-specific analysis of the connection between the model optimization goal (forecasting) and feature-oriented physics. Fig.~\ref{fig:topk_feature_specific}(b) shows how each feature is associated with its own projection vector $\mathbf{p}$, leading to feature‑specific masks, whereas Fig.~\ref{fig:topk_feature_specific}(c) provides an illustrative snapshot from the SPEEDY dataset (described in Sec.~\ref{sec:datasets} below), displaying the resulting feature-specific masks for temperature $T$, specific humidity $q$, and the 500\,hPa wind components $u_{500}$ and $v_{500}$.

To encourage meaningful interpretability, the loss function is augmented with a regularization term that explicitly ties the importance masks to prediction error. The total loss is defined as:

\begin{equation}
    \mathcal{L} = \text{MSE}(x_{\text{pred}}, x_{\text{target}}) + \lambda \sum_{f=1}^{N_F} \frac{1}{\text{Budget}_f},
    \label{eq:total_loss}
\end{equation}

where the first term is the standard mean squared error, and the second term penalizes low error budget attribution in the sub-sampled nodes for each feature. The budget for feature \( f \) is computed as:

\begin{equation}
    \text{Budget}_f = \text{MSE}(m_f \odot x_{\text{pred}, f}, m_f \odot x_{\text{target}, f}),
    \label{eq:budget}
\end{equation}

where \( m_f \in \{0,1\}^N \) is the binary mask (a masked field) for feature \( f \) and \( \odot \) indicates element-wise multiplication. This formulation ensures that the selected nodes correspond to regions where the feature-specific prediction error is concentrated.

Although the interpretability module appreciably enlarges the model's parameter space, this additional complexity yields substantive benefits. The supplementary layers enrich the learned latent spaces, enabling more expressive feature representations that capture multiscale structure with higher fidelity. More importantly, the masking mechanism renders the decision process transparent: it pinpoints the spatial and feature‐level contributions that govern each prediction, facilitating rigorous diagnostic analysis, guiding mesh or sensor refinement, and revealing potential physics‐based insights. 


\subsection{Dataset Descriptions}
\label{sec:datasets}
\begin{figure*}
    \centering
    \includegraphics[width=\textwidth]{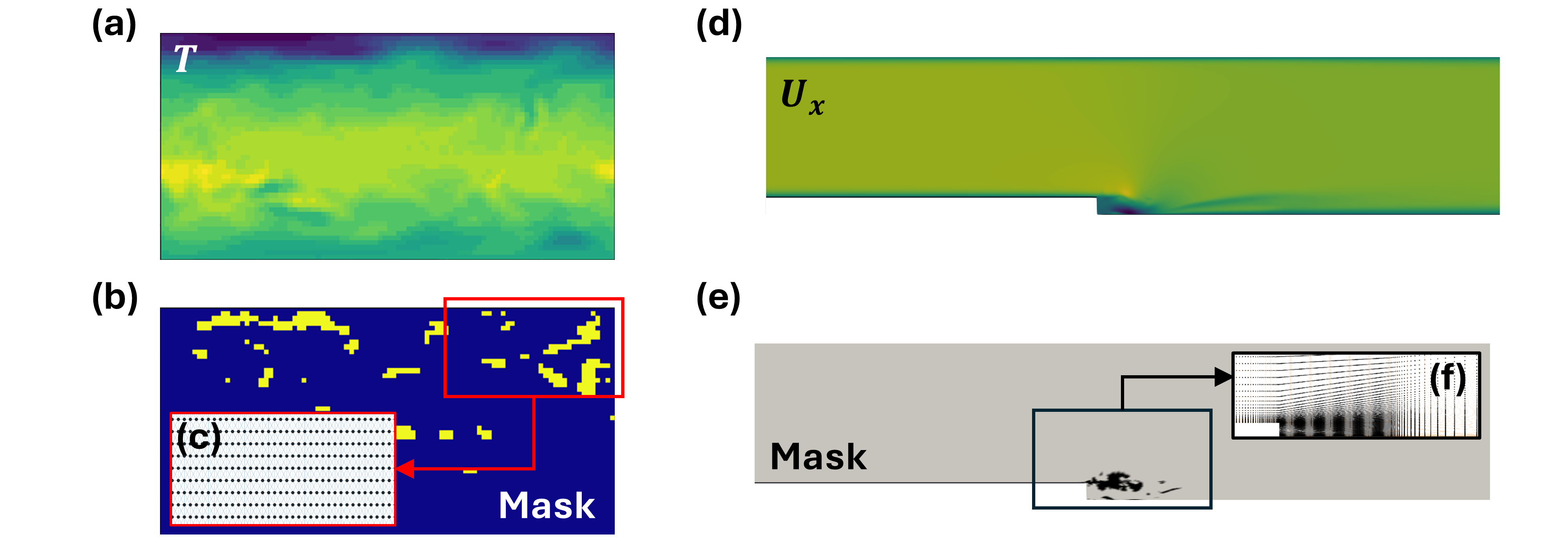}
    \caption{%
       Feature‑specific Top‑$K$ masks on the two benchmark datasets.
       (a) Temperature field $T$ from the SPEEDY dataset at time $t=t_0$.
       (b) Temperature mask produced by FIGNN at the same instant; yellow pixels mark nodes retained by the Top‑$K$ operator.
       (c) Enlargement of the red rectangle in~(b), revealing the regular latitude–longitude grid and the edges that connect neighbouring nodes.  
       (d) Stream‑wise velocity $U_x$ for a snapshot from the BFS dataset.
       (e) Corresponding Top‑$K$ mask for $U_x$ on the unstructured CFD mesh; the grey region is the domain, black points indicate the retained sub‑graph.  
       (f) Close‑up of the boxed step region in~(e) showing individual cell centroids (nodes) and their finite‑volume connectivity.  
       Together these panels illustrate that, despite the very different spatial discretisations (regular Eulerian grid vs.\ unstructured control‑volume mesh), FIGNN consistently extracts sparse, task‑relevant sub‑graphs that concentrate around dynamically important flow structures.}
    \label{fig:mask_examples}
\end{figure*}

FIGNN is demonstrated in this work to model fluid flows in two physically distinct and complex configurations using the above-described surrogate modeling training setup. The first demonstration case involves climate dynamics prediction; the training dataset for this task is sourced from the SPEEDY dataset, which captures large-scale, real-world climate dynamics data on a two-dimensional grid. The second case involves predicting high-Reynolds-number separated flow over a backward-facing step (BFS) geometry, with the target data produced using numerical simulations in OpenFOAM. These two problem domains are each highly complex, yet differ significantly physical flow structure, providing a rigorous testbed for assessing the generality and effectiveness of the proposed interpretability framework. Specific details for each are described below.

\textbf{Climate dynamics:} The climate dynamics dataset used in this work -- termed the SPEEDY dataset -- is derived from a simplified global circulation model that outputs multiple atmospheric variables over a latitude--longitude grid \cite{molteni2003atmospheric,kucharski2006decadal}. Each grid cell corresponds to a node \(i \in \mathcal{V}\), and an edge \((i,j)\) is formed whenever the corresponding cells are adjacent on the spherical surface. The SPEEDY grid has a spatial resolution of $3.75^{\circ} \times 3.75^{\circ}$ -- in this work, for each node, the feature vector includes:
\[
\bigl(T,\, q,\, u_{500},\, v_{500}\bigr),
\]
where \(T\) denotes temperature, \(q\) denotes specific humidity, and \(u_{500}, v_{500}\) denote the horizontal wind components at the 500-hPa level. This multi-variate representation allows a graph-based model to capture both local and global atmospheric interactions. The dataset is particularly useful for assessing feature-level interpretability because each variable exerts distinct influence on the global circulation, thereby testing the ability of the interpretability framework to isolate those influences across different regions of the globe. A snapshot of $T$ is shown in Fig.~\ref{fig:mask_examples}(a), with its corresponding masked field shown in Fig.~\ref{fig:mask_examples}(b). A zoomed-in plot Fig.~\ref{fig:mask_examples}(c) shows the underlining graph connectivity. The training and testing sets are comprised of 7,442 and 826 graphs respectively, with each graph consisting of 4,608 nodes and 46,976 edges. Nodes correspond to latitude-longitude grid points mapped to planar $(x,y)$ coordinates, and edges are constructed using $k=10$ Euclidean nearest node neighbors (bidirectional edges), capturing regional advection patterns while maintaining sparse connectivity. For additional dataset details, the reader is directed to Refs.~\cite{molteni2003atmospheric,kucharski2006decadal}. The baseline architecture contains 1,968,388 trainable parameters, whereas the FIGNN model scales this up to 14,541,312 parameters.

\textbf{Backward-facing step flow:} The Backward-Facing Step (BFS) dataset is sourced from two-dimensional OpenFOAM simulations, as described in previous works \cite{barwey2023multiscale,barwey2025interpretable}. The spatial domain is discretized into a mesh whose cell centroids form the graph node set \(\mathcal{V}\). Two nodes are connected by an edge if they share a cell face, reproducing the finite-volume stencil (with bidirectional edges $(i,j)$ and $(j,i)$ explicitly stored). This preserves native connectivity patterns required for flux calculations, aligning with mesh-based GNN strategies for CFD. This particular dataset, ranging from $Re = 26,212$ to $45,589$, contains the two-component velocity field $\bigl(u_{x},\, u_{y}\bigr) $, where \(u_{x}\) and \(u_{y}\) denote the horizontal and vertical velocity components at each node. The training and testing sets contain 7,980 and 401 graphs, respectively, with each graph having 20,540 nodes and 81,494 edges. Fig.~\ref{fig:mask_examples}(d) shows a snapshot of $U_x$, with its corresponding mask in Fig.~\ref{fig:mask_examples}(e), while Fig.~\ref{fig:mask_examples} show a zoomed view of the mesh near the step region, demonstrating the cell centroids and connectivity. The BFS dataset showcases strong flow gradients near the step, at characteristic separation-induced shear layer, in step-cavity recirculation zone, and at the downstream flow reattachment point, making it an ideal testbed for feature-specific interpretability. A graph-based model can exploit local connectivity to forecast or approximate flow patterns, and an interpretability framework can highlight the relative importance of \(u_{x}\) versus \(u_{y}\) in different regions (for instance, in the recirculation zone or along the shear layer). For further detail on BFS dataset generation, the reader is directed to Ref.~\cite{barwey2025interpretable}. For this dataset, the baseline architecture comprises 1,967,874 trainable parameters, whereas the feature-specific FIGNN expands the count to 7,270,656.

Demonstrations using both SPEEDY and BFS datasets provides a sufficiently broad evaluation scope, as these are fundamentally different physical systems. SPEEDY captures large-scale geophysical fluid dynamics, with multiple atmospheric variables influencing highly spatially-distributed dynamics, whereas BFS offers a canonical separated flow testbed with more spatially coherent unsteady flow patterns. As a result, the goal of the section below is to showcase feature-level interpretability in both settings to  underscore the generality of the interpretability approach introduced here, showing that it can disentangle the contributions of individual variables in markedly different physical domains.

\section{Results and Discussion}

\begin{figure}
    \centering
    \includegraphics[width=1\textwidth]{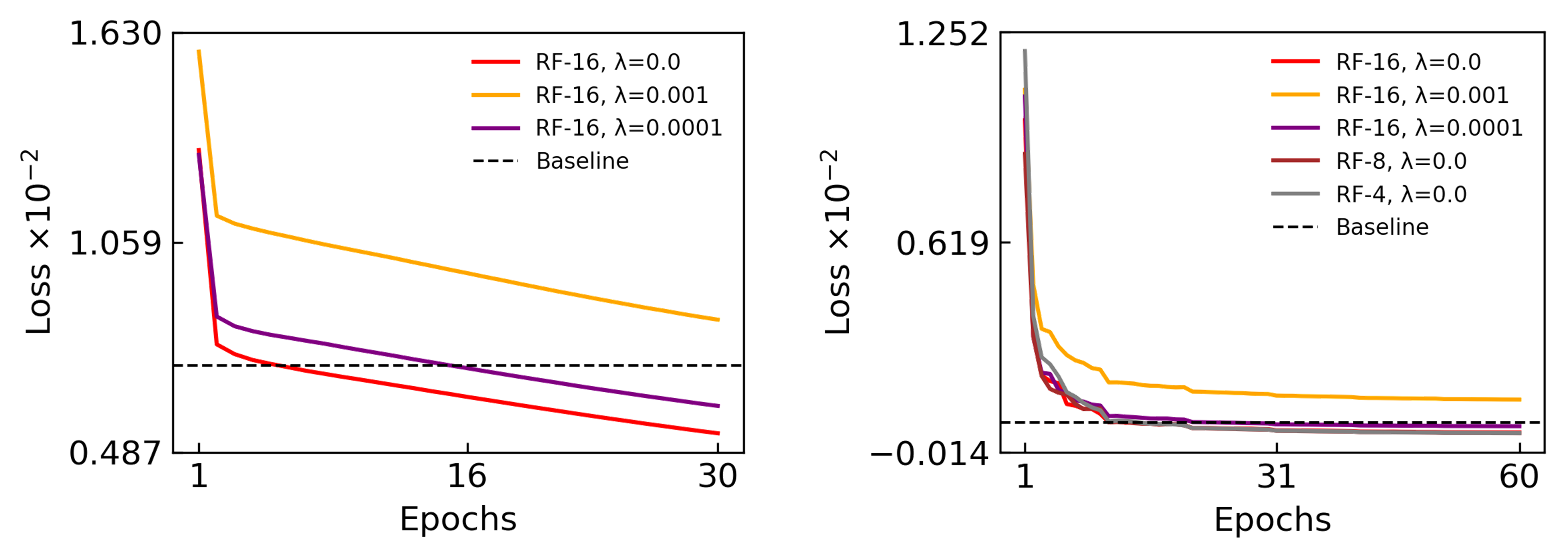}
    
    
    \caption{%
    Training loss curves for interpretable (solid lines) versus baseline (dashed line) GNNs 
    on the SPEEDY dataset (left) and the BFS dataset (right). Models without significant 
    interpretability regularization converge to lower or comparable loss values. 
    Higher regularization increases interpretability at the cost of elevated 
    final loss.%
    }
    \label{fig:loss_curves}
\end{figure}

Figure~\ref{fig:loss_curves} compares the training loss for FIGNNs (solid lines) against a baseline GNN model (dashed line) on two datasets: SPEEDY (left) and BFS (right). In both cases, FIGNN models with minimal or no regularization converge to losses that are equal to or lower than those of the baseline, indicating that selectively focusing on relevant features does not necessarily impede accuracy. By contrast, higher regularization intensifies the interpretability objective but typically raises the final loss, reflecting a trade-off between predictive fidelity and clarity regarding feature importance. The BFS task converges more rapidly overall, likely due to its lower dimensional feature space (\(u_x, u_y\)) compared to the broader set of atmospheric variables (\(T,\,q,\,u_{500},\,v_{500}\)) in SPEEDY. Despite these differences, both datasets exhibit a similar balance between interpretability and performance when adjusting regularization strength, similar to previously observed budget regularization trends \cite{barwey2025interpretable}.
\subsection{SPEEDY Dataset}
\begin{figure}
    \centering
    \includegraphics[width=\textwidth]{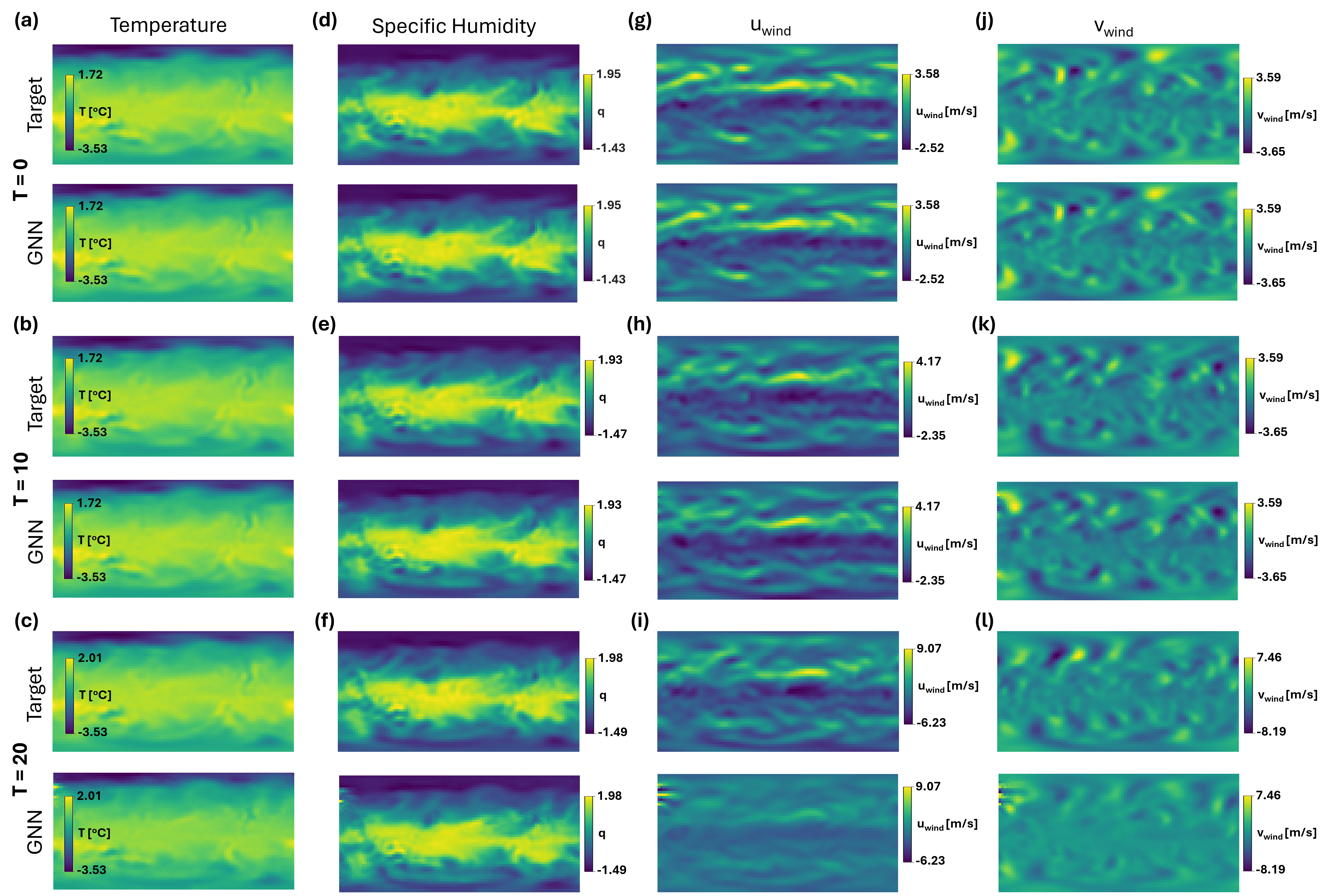}
    \caption{Multi-step rollout predictions for each atmospheric feature from the SPEEDY dataset, produced by FIGNN (RF = 16, $\lambda = 10^{-4}$). Each panel displays the temporal evolution of a single feature—temperature \(T\), specific humidity \(q\), zonal wind \(u_{500}\), and meridional wind \(v_{500}\).}
    \label{fig:feature_rollout}
\end{figure}

Figure~\ref{fig:feature_rollout} shows the multi-step rollout predictions for each individual atmospheric feature from the SPEEDY dataset, generated by the FIGNN–RF16 model trained with a budget-regularisation coefficient $\lambda = 10^{-4}$. In this figure, separate panels illustrate the temporal evolution of temperature (\(T\)), specific humidity (\(q\)), zonal wind (\(u_{500}\)), and meridional wind (\(v_{500}\)). The rollout is organized by forecast steps, demonstrating that FIGNN is capable of producing coherent and stable predictions for each feature as the forecast progresses. The clear progression in each panel underscores that the underlying spatiotemporal relationships in the atmospheric data are effectively captured, even when extending the prediction over multiple time steps. However, over the length of the rollout, errors keep building up since predictions are fed back to the model as inputs for subsequent predictions. This results in a deterioration of the model's predictive capabilities over long time horizons, as seen by the blurred predictions of $u_{500}$ and $v_{500}$ fields in the $20^{th}$ timestep. It is emphasized that this form of error accumulation is largely a by-product of single-step data-based training, and is not tied to the FIGNN framework - approaches with enhanced long-time prediction (either through incorporation of physical constraints or rollout-based training) can readily be integrated. 

\begin{figure}
    \centering
    \includegraphics[width=\textwidth]{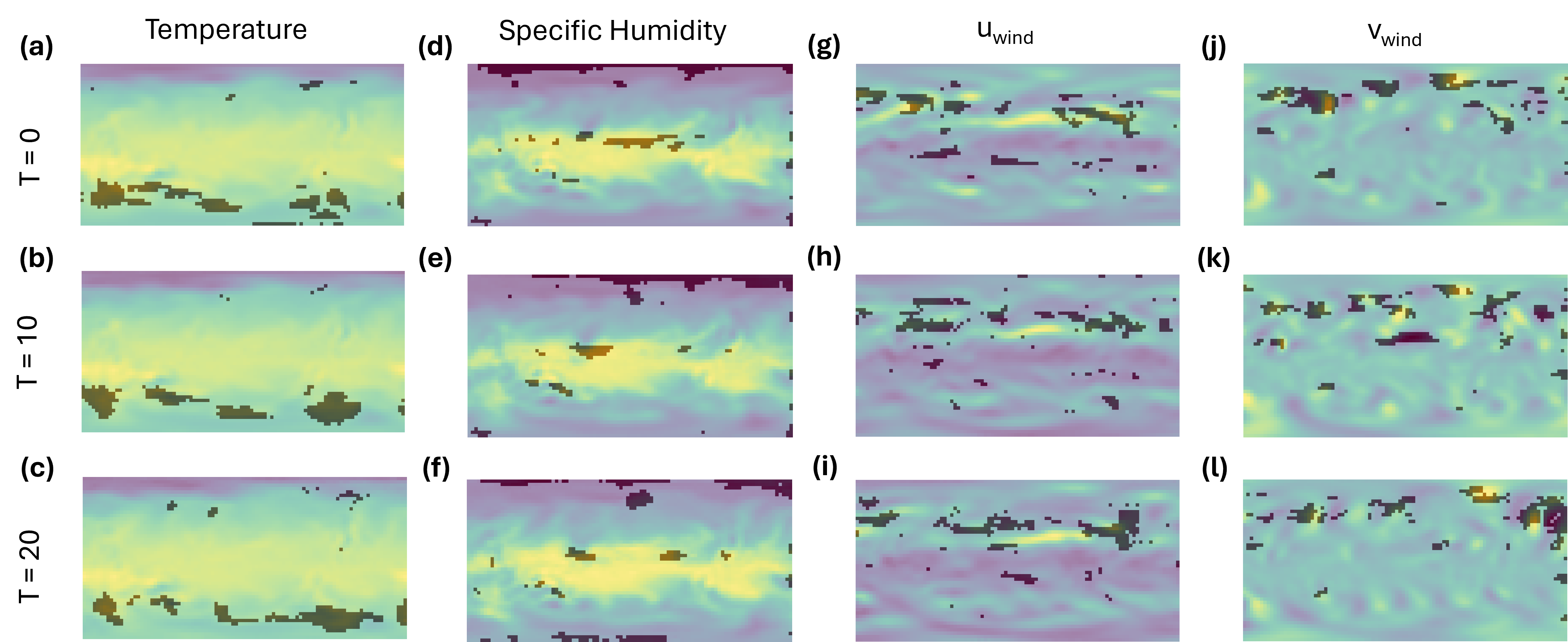}
    \caption{Overlay of feature-specific interpretability masks on the multi-step rollouts for the SPEEDY dataset. In each panel, the mask highlights the spatial regions that FIGNN deems most critical for predicting the corresponding feature (e.g., \(T\), \(p\), \(u_{500}\), \(v_{500}\)). Brighter regions indicate nodes or grid cells with higher importance, thereby enhancing interpretability by isolating feature-specific structures that drive the model's predictions.}
    \label{fig:feature_mask_overlay}
\end{figure}

Figure~\ref{fig:feature_mask_overlay} builds upon these predictions by overlaying feature-specific interpretability masks on the same multi-step rollouts. In these overlays, the feature-specific masks—generated through independent TopK pooling branches for each feature—highlight the spatial regions that are deemed most critical for predicting the corresponding variable. For example, in the temperature panel, the mask isolates regions that likely correspond to significant thermal gradients or frontal boundaries; in the wind panels (\(u_{500}\) and \(v_{500}\)), the masks emphasize areas with pronounced flow characteristics. Brighter regions indicate nodes or grid cells with a higher importance score, while darker regions denote lower relevance. It is emphasized that this visualization directly links the model’s internal feature attribution to the corresponding physical structures, thereby enhancing interpretability on a feature-specific level. In other words, the masks provide a connection between the model prediction goal (forecasting), architecture (multiscale message passing), and spatially coherent structures relevant to each feature in the prediction process. These feature-specific masks can then be connected to domain-specific physical phenomena in a user-guided analysis step, providing futher granularity in the interpretability process over the single-mask counterpart \cite{barwey2025interpretable}. 

Together, these figures illustrate the dual benefit of the FIGNN framework for the SPEEDY dataset. Crucially, the above figures illustrate that the feature-specific masks provide detailed, spatially resolved insight into which features contribute most significantly to the forecast, \textit{while retaining} the surrogate GNN forecasting workflow of predicting the evolution of individual features over time.

\begin{figure}
    \centering
    \includegraphics[width=\linewidth]{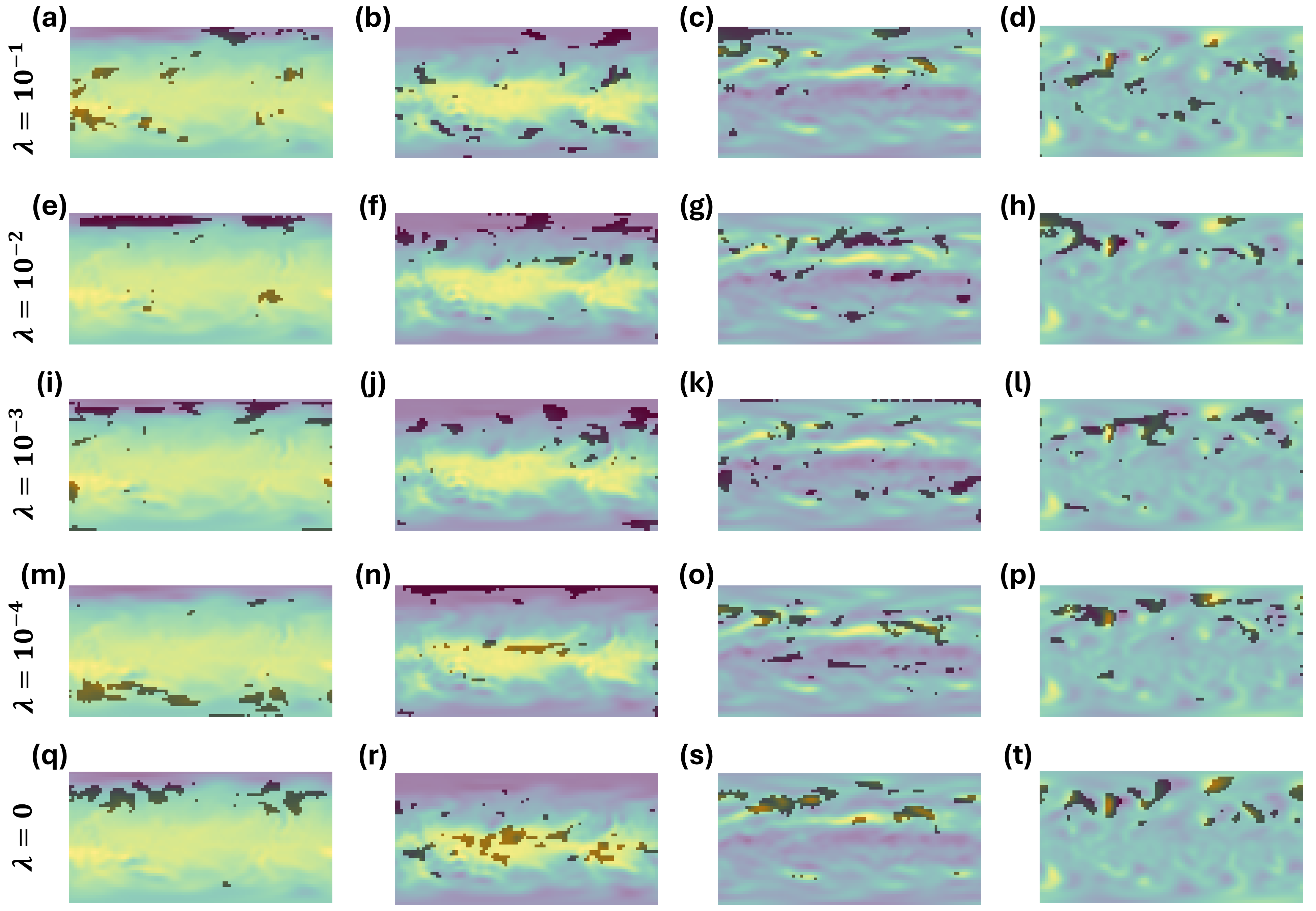}
    \caption{%
        Effect of the error--tagging coefficient $\lambda$ on feature-specific Top-K masks (RF = $16$).
        Each row shows the fixed--size mask for four atmospheric features, while sweeping $\lambda$ from $10^{-1}$ (top) to $0$ (bottom).%
        Because the reduction factor pins the mask cardinality, increasing $\lambda$ only re-orders \emph{which} nodes are retained: large $\lambda$ concentrates on maximising the percentage of MSE captured within the mask, yielding fragmented, spatially incoherent masks; moderate to low $\lambda$ trades some budget for contiguity, producing clusters that align with physically meaningful structures (e.g.\ jet cores, boundary layers), and thereby offers greater interpretability.%
    }
    \label{fig:lambda_mask_sweep}
\end{figure}

Figure~\ref{fig:lambda_mask_sweep} sweeps the error–tagging coefficient $\lambda$ from $10^{-1}$ to~$0$ while the reduction factor is held fixed at $\text{RF}=16$.  Because the mask size is immutable, changes in the shaded regions arise solely from $\lambda$ re-ordering which nodes are selected for those $K$ slots; panels within each row display the same instantaneous state of the SPEEDY atmospheric variables—temperature, specific humidity, and the zonal and meridional wind components—so any spatial shift can be interpreted in physical context.

At the largest values of $\lambda$ ($10^{-1}$ and $10^{-2}$) the masks appear as highly fragmented structures, scattered across broad swaths of the domain.  Under this strong penalty the optimiser maximises the fraction of single-step mean-squared error captured inside the fixed mask and therefore sprinkles isolated nodes over many high-error outliers, sacrificing spatial coherence.

Reducing $\lambda$ to $10^{-3}$ and $10^{-4}$ causes the same number of nodes to coalesce into elongated bands that trace the mid-latitude jet, tropical convection zones, and boundary-layer shear fronts.  In this regime the optimiser tolerates a modest drop in error coverage to gain contiguity, yielding masks that align with physically recognisable features and thus offer greater interpretability.

With $\lambda$ set to zero, the error-budget term is absent and the nodes are selected solely for the purpose of global MSE minimization. Because the baseline weights are frozen during fine-tuning, these nodes simply trace the network’s intrinsic predictive attention, confirming that the regulariser—rather than the pooling layer itself—is responsible for steering the mask toward regions of high residual.

Overall, large $\lambda$ values maximise error coverage but fragment the mask, whereas small $\lambda$ values improve spatial coherence at the cost of leaving more residual outside the mask.  For the SPEEDY dataset examined here, intermediate values in the range $10^{-3}$–$10^{-4}$ appear to strike the best balance between quantitative budget and qualitative interpretability.

\begin{figure}
    \centering
    \includegraphics[width=1\textwidth]{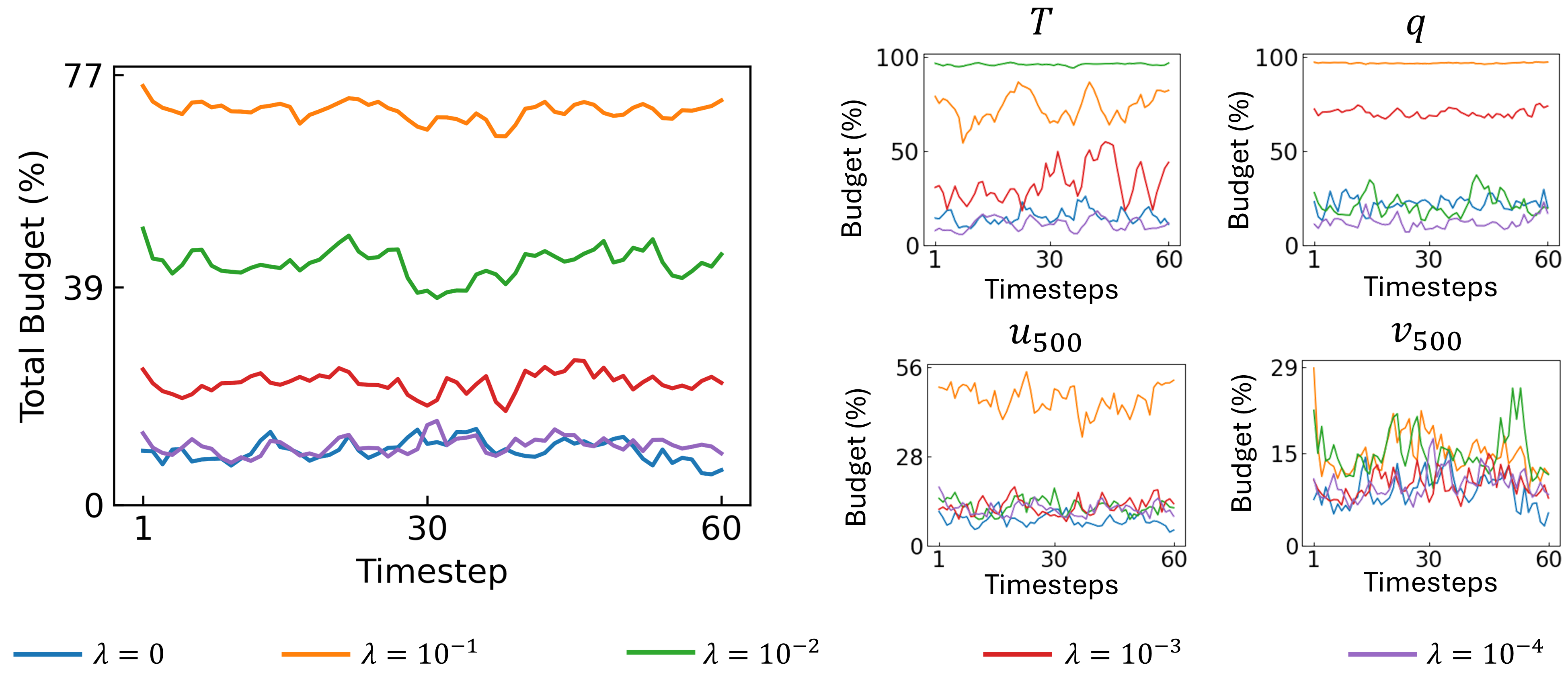}
    \caption{Single-step budget plots for the FIGNN model on the SPEEDY dataset. The left panel displays the total error budget (i.e., the fraction of overall forecasting error that is confined to the nodes selected by the feature-specific masks) as a function of different regularization parameters \(\lambda\). The right panels show the individual contributions from each feature ($T$, $q$, \(u_{500}\), and \(v_{500}\)) under the same conditions.}
    \label{fig:budget_single_step}
\end{figure}

\begin{figure}
    \centering
    \includegraphics[width=1\textwidth]{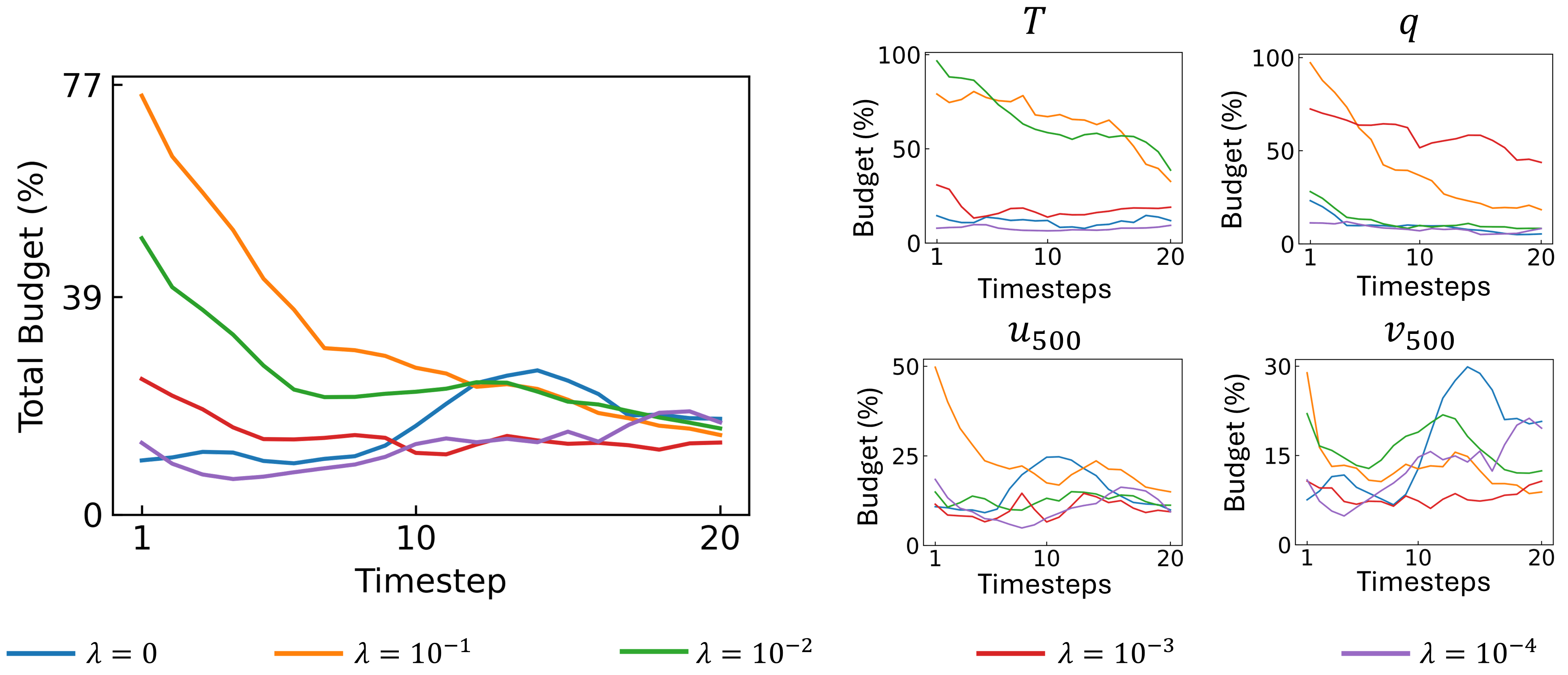}
    \caption{Rollout budget plots for the FIGNN model on the SPEEDY dataset. This figure illustrates the evolution of the total budget over multiple forecast steps for varying \(\lambda\) values. Although higher \(\lambda\) settings concentrate a larger fraction of the forecasting error within the masked regions during the early rollout steps, this effect gradually diminishes over longer forecast horizons due to the accumulation and diffusion of error.}
    \label{fig:budget_rollout}
\end{figure}

Figure~\ref{fig:budget_single_step} presents the budget evaluation results in a single-step prediction scenario. The left panel reports the total error budget, which quantifies the fraction of the overall forecasting error that is confined within the nodes selected by the feature-specific masks. The individual panels on the right decompose this error budget into separate contributions from each feature—namely, $T$, $q$, \(u_{500}\), and \(v_{500}\). Under higher regularization settings (i.e., larger \(\lambda\)), the total budget is increased, indicating that a larger proportion of the prediction error is being localized within the masked subgraphs. In contrast, lower values of \(\lambda\) yield smaller budget percentages, as there is less incentive for the model to concentrate error into the masked regions. Notably, the feature-specific breakdown reveals that certain features consistently capture a greater share of the error, suggesting that the model naturally prioritizes error tagging in variables with larger or more dynamic prediction discrepancies.

Figure~\ref{fig:budget_rollout} illustrates the evolution of the error budget during multi-step (rollout) forecasts. The plots show that, when the regularization strength is high, the total budget is initially elevated—implying effective error localization by the masks—in the early rollout steps. However, as the forecast horizon extends, the ability to confine error within the masked regions diminishes. This trend is attributed to the accumulation and diffusion of errors over successive prediction steps, which makes it more challenging for the model to consistently tag errors. Conversely, when \(\lambda\) is set to a lower value, the rollout budget remains more stable over time, albeit at lower levels of error tagging.

Together, these budget analyses reveal a fundamental trade-off: higher regularization \(\lambda\) enhances interpretability by forcing the model to isolate a greater fraction of the error within feature-specific masks, yet it may also result in a slight degradation of overall forecast accuracy. Lower \(\lambda\) values, while preserving prediction fidelity closer to a non-regularized baseline, provide a less pronounced error budget. 

\subsection{BFS Dataset}

\begin{figure}
    \centering
    \includegraphics[width=\textwidth]{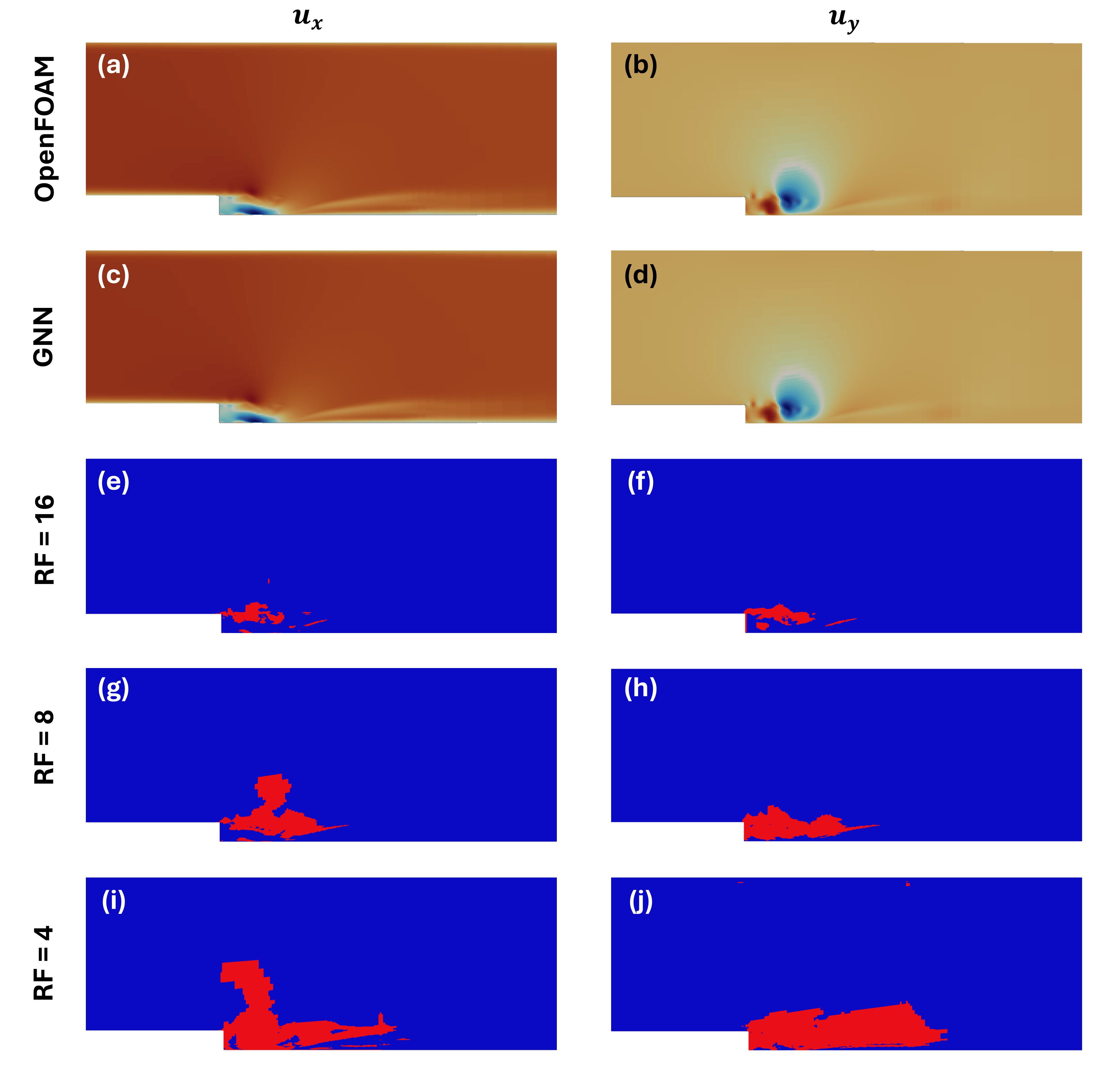}
    \caption{Comparison of FIGNN predictions for the backward-facing step (BFS) dataset. In the top panels, the surrogate model’s predicted velocity fields (streamwise \(u_x\) and vertical \(u_y\)) are compared against the high-fidelity CFD (OpenFOAM) ground-truth. The lower panels display feature-specific interpretability masks generated using different node reduction factors (RF). Here, RF=16, being the most restrictive, selects the fewest nodes and produces a sparse mask that concentrates on critical flow regions (e.g., near the step and reattachment zones). In contrast, lower RF values (RF=8 and RF=4) generate denser masks covering broader areas of the domain.}
    \label{fig:BFS_RF}
\end{figure}

Figure~\ref{fig:BFS_RF} presents the performance of FIGNN on the BFS dataset. In the top panels, the surrogate model’s predicted velocity fields for both the streamwise component (\(u_x\)) and the vertical component (\(u_y\)) are compared against the high-fidelity CFD (OpenFOAM) ground-truth. The predicted flow structures capture the key aerodynamic features—including flow separation at the step, recirculation zones, and reattachment dynamics—with only minor discrepancies observed in the vortex-shedding regions.

The lower panels illustrate the feature-specific interpretability masks obtained using different node reduction factors (RF). A higher reduction factor (RF=16) is the most restrictive, meaning that it selects fewer nodes for masking, which produces a sparser mask that concentrates on the most critical regions of the flow (e.g., near the step and in the reattachment zones). In contrast, lower reduction factors (RF=8 and RF=4) retain more nodes, generating denser masks that provide a broader coverage of the domain. This variation in mask sparsity not only confirms that the FIGNN framework can accurately reproduce the flow field but also demonstrates its capability to provide insight into which localized regions and features are most influential for the model’s predictions.
\begin{figure}[ht]
    \centering
    \includegraphics[width=\linewidth]{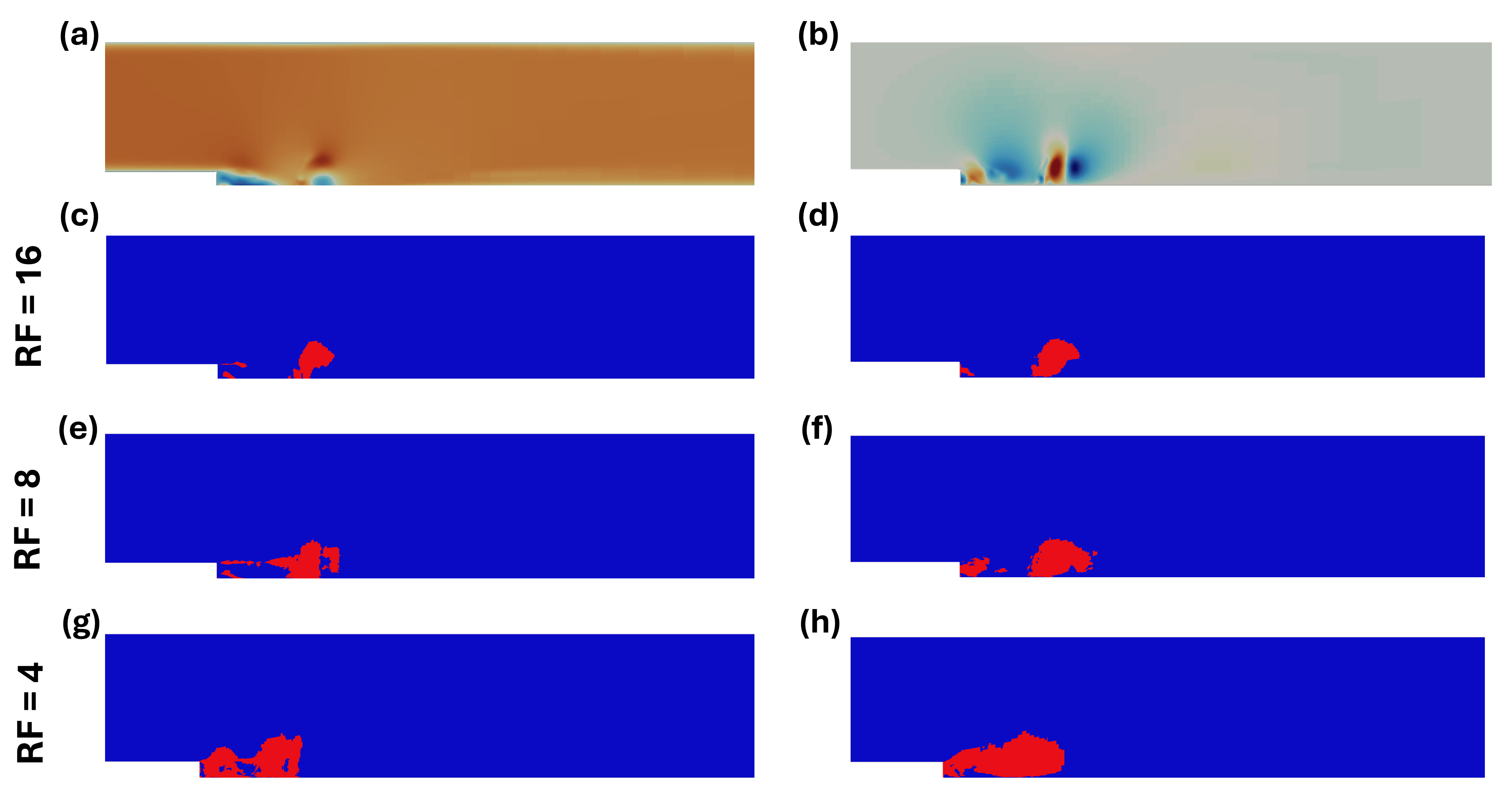}
    \caption{%
        Full-field predictions and feature-specific Top-K masks for the backward–facing-step flow at timestep~60 (Re = $26,212$).
        Panels~(a) and~(b) show the model predictions for the instantaneous stream-wise velocity $u_x$ and cross-stream velocity $u_y$, respectively.%
        Rows~2–4 overlay the corresponding error-tagging masks (red) for three reduction factors, $\mathrm{RF}=16$ (c,d), $8$ (e,f), and $4$ (g,h); blue nodes are outside the mask.%
        For every RF both velocity components focus exclusively on the separated shear-layer region downstream of the step, where flow reversal and vortex roll-up dominate the dynamics.%
        Increasing the node budget (smaller RF) does not shift the mask elsewhere; it simply enlarges the tagged area within the same separation bubble—along the reversed-flow core for $u_x$ and along successive shed vortices for $u_y$.%
    }
    \label{fig:bfs_mask_rf_sweep}
\end{figure}

Figure~\ref{fig:bfs_mask_rf_sweep} examines the backward-facing-step simulation at timestep 60 ($\mathrm{Re}=26,212$), a moment showcasing downstream vortex shedding and flow re-attachment. Panels (a) and (b) show the network’s predictions for the stream-wise velocity $u_x$ and the cross-stream velocity $u_y$, respectively. The remaining panels superimpose feature-specific Top-K masks for three node budgets, achieved by setting the reduction factor to $\text{RF}=16,\,8,$ and 4—corresponding to $|V|/16,\;|V|/8,$ and $|V|/4$ retained nodes.

Across every budget, and for both velocity components, the mask remains anchored to the separation bubble just downstream of the step. No nodes are allocated to the channel core or to the smooth upper wall, indicating that the network consistently identifies the separated region as the main source of predictive difficulty at this instant.

The two features nevertheless highlight complementary aspects of the same mechanism. For $u_x$ the smallest mask ($\text{RF}=16$) selects a compact patch centered on the pocket of reversed stream-wise flow, where velocity gradients are steepest. Increasing the node budget ($\text{RF}=8$ and 4) causes the patch to thicken and extend along the shear layer, eventually covering the entire separation bubble and its downstream re-attachment zone. In contrast, the $u_y$ masks track the cross-stream streaks associated with vortex roll-up: even with the tightest budget the selected nodes sit just above the bottom wall inside the first vortex core, and additional capacity elongates the patch through successive vortices without drifting laterally or upstream.

This behaviour shows that enlarging the mask does not divert attention to new, less-relevant areas; instead, the extra nodes are invested locally to refine coverage of the region where unsteady shear, flow reversal, and vortex shedding interact. The coherence and stability of the masks across budgets—and their agreement with established flow-physics expectations—provide strong evidence that the error-tagging strategy delivers physically meaningful interpretability while allowing precise control over the number of nodes inspected.

\begin{figure}
    \centering
    \includegraphics[width=1\textwidth]{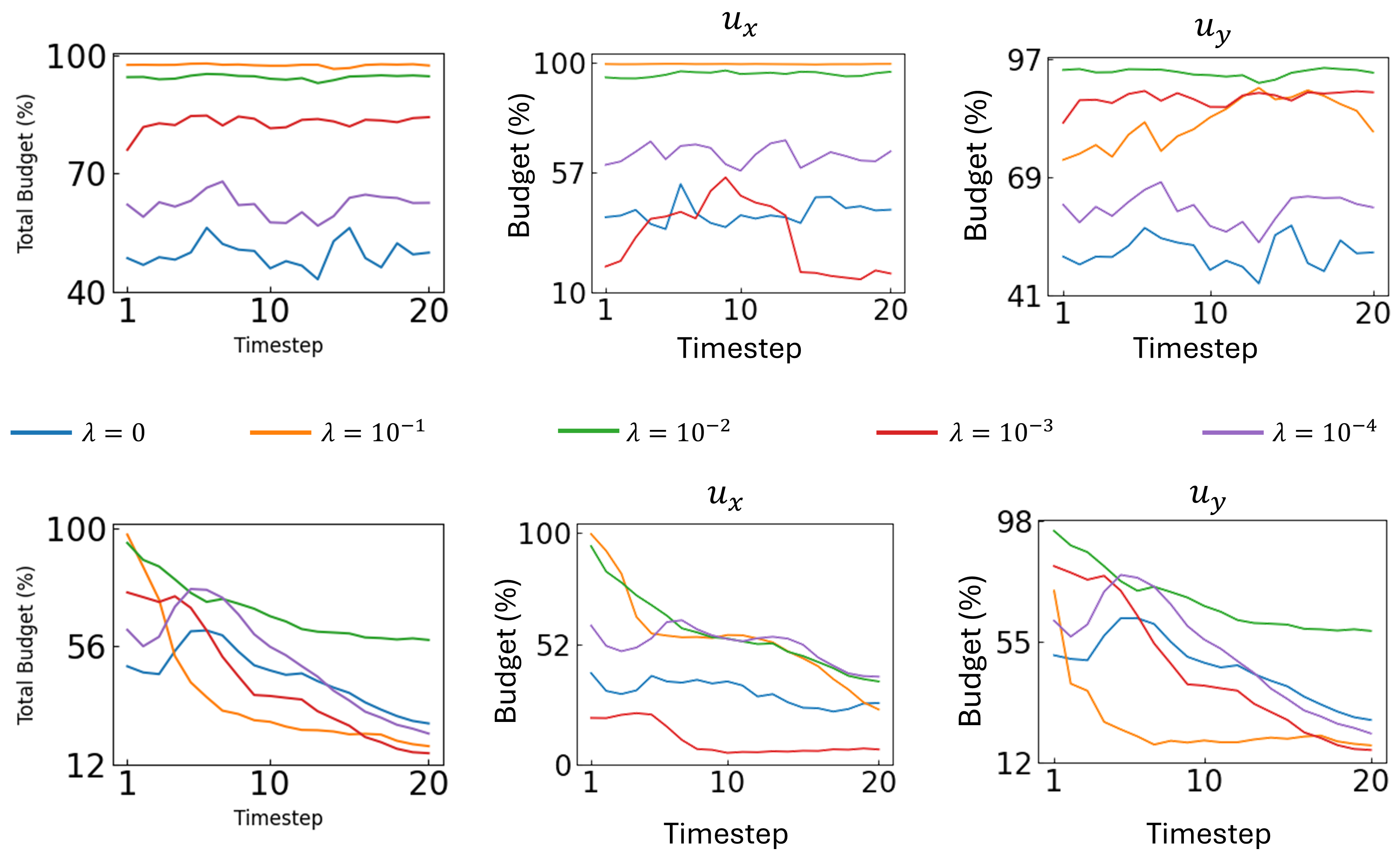}
    \caption{Budget plots for the backward-facing step (BFS) dataset. The top row displays the single-step budgets, indicating the fraction of the prediction error confined within the masked regions for both the streamwise (\(u_x\)) and vertical (\(u_y\)) velocity components under varying regularization parameters (\(\lambda\)). The bottom row shows the rollout budgets that capture how this error localization evolves over multiple prediction steps. Higher \(\lambda\)-values result in masks that concentrate a larger portion of the error in the short term, whereas over longer forecasting horizons, the advantage of high \(\lambda\) diminishes as prediction errors accumulate.}
    \label{fig:BFS_budget}
\end{figure}

Figure~\ref{fig:BFS_budget} presents the error budget analysis for the backward-facing step (BFS) dataset under different regularization settings. The top row illustrates the single-step budgets, which quantify the fraction of the overall prediction error that is confined within the masked regions for each velocity component. Here, higher \(\lambda\)-values concentrate more of the single-step error within the masked nodes, indicating a stronger emphasis on error localization through the feature-specific masks. In contrast, lower \(\lambda\) values or the absence of regularization yield lower budget values, as the model is not explicitly penalized for spreading errors outside the masked regions.

The bottom row displays the rollout budgets, showing how the error localization evolves over multiple prediction steps. In the rollout scenario, models initialized with high \(\lambda\) start with a larger fraction of error being localized in the masked regions at the early steps. However, as the forecast horizon extends, the advantage of concentrating error in the masks diminishes due to the cumulative propagation and diffusion of errors through the unmasked regions. This behavior suggests a trade-off analogous to what was observed in the SPEEDY problem: while higher \(\lambda\) enhances short-term interpretability by tightly focusing the errors, it may also lead to a reduction in the long-term stability of error confinement.

\section{Conclusions}

This study introduced a feature-specific interpretable Graph Neural Network (FIGNN) framework to predict complex spatio-temporal dynamics in atmospheric (SPEEDY) and fluid-flow (Backward-Facing Step, BFS) datasets. By employing multiple feature-specific TopK masks, FIGNN successfully identified distinct regions and variables critical to accurate predictions, significantly enhancing interpretability compared to previous single-mask approaches.

On the SPEEDY dataset, the model demonstrated stable and coherent predictions across multiple forecast steps for each atmospheric variable (temperature, specific humidity, and wind components). Feature-specific masks highlighted physically meaningful structures, such as frontal boundaries and prominent flow regions, thereby confirming alignment between the model’s learned features and established atmospheric dynamics. Additionally, the budget analyses revealed that stronger interpretability regularization effectively localized prediction errors within masked regions, albeit with a slight trade-off in long-term predictive stability.

Similarly, on the BFS dataset, FIGNN accurately reproduced essential fluid dynamics phenomena, including separation, recirculation, and reattachment regions. Feature-specific masks identified critical areas around the step and within recirculation zones, and varying the node reduction factor (RF) allowed controlled exploration of the balance between mask sparsity and interpretability. Budget analyses reinforced that higher regularization localized errors more effectively in the short term but demonstrated diminishing returns over extended forecast horizons due to cumulative error propagation.

Overall, the presented FIGNN approach provides a balance between interpretability and predictive performance, offering domain experts detailed insights into feature-specific contributions to prediction accuracy. Future work may explore extending this framework to spatially three-dimensional settings, real-time adaptive interpretability, and uncertainty quantification strategies to further broaden its applicability in scientific modeling. Additionally, the structures revealed by FIGNN may be further analysed for their utility in improving the solution of inverse problems arising from super-resolution, control, optimization, etc.

\section{Acknowledgments} 
This research used resources of the Argonne Leadership Computing Facility, which is a U.S. Department of Energy Office of Science User Facility operated under contract DE-AC02-06CH11357. SB acknowledges support by the U.S. Department of Energy (DOE), Office of Science under contract DE-AC02-06CH11357. RM acknowledges funding support from ASCR for DOE-FOA-2493 ``Data-intensive scientific machine learning'' and from the RAPIDS2 Scientific Discovery through Advanced Computing (SciDAC) program. RM and RR acknowledge computational support from Penn State Institute for Computational and Data Sciences.

\bibliographystyle{elsarticle-num}
\bibliography{references}

\begin{thebibliography}{10}
\expandafter\ifx\csname url\endcsname\relax
  \def\url#1{\texttt{#1}}\fi
\expandafter\ifx\csname urlprefix\endcsname\relax\def\urlprefix{URL }\fi
\expandafter\ifx\csname href\endcsname\relax
  \def\href#1#2{#2} \def\path#1{#1}\fi

\bibitem{li2020multipole}
Z.~Li, N.~Kovachki, K.~Azizzadenesheli, B.~Liu, A.~Stuart, K.~Bhattacharya, A.~Anandkumar, Multipole graph neural operator for parametric partial differential equations, Advances in Neural Information Processing Systems 33 (2020) 6755--6766.

\bibitem{wu2022graph}
L.~Wu, P.~Cui, J.~Pei, L.~Zhao, X.~Guo, Graph neural networks: foundation, frontiers and applications, in: Proceedings of the 28th ACM SIGKDD conference on knowledge discovery and data mining, 2022, pp. 4840--4841.

\bibitem{bronstein2017geometric}
M.~M. Bronstein, J.~Bruna, Y.~LeCun, A.~Szlam, P.~Vandergheynst, Geometric deep learning: going beyond euclidean data, IEEE Signal Processing Magazine 34~(4) (2017) 18--42.

\bibitem{kurz2025harnessing}
M.~Kurz, A.~Beck, B.~Sanderse, Harnessing equivariance: Modeling turbulence with graph neural networks, arXiv preprint arXiv:2504.07741 (2025).

\bibitem{horie2022physics}
M.~Horie, N.~Mitsume, Physics-embedded neural networks: Graph neural pde solvers with mixed boundary conditions, Advances in Neural Information Processing Systems 35 (2022) 23218--23229.

\bibitem{han2022predicting}
X.~Han, H.~Gao, T.~Pfaff, J.-X. Wang, L.-P. Liu, Predicting physics in mesh-reduced space with temporal attention, arXiv preprint arXiv:2201.09113 (2022).

\bibitem{gilmer}
J.~Gilmer, S.~S. Schoenholz, P.~F. Riley, O.~Vinyals, G.~E. Dahl, Neural message passing for quantum chemistry, in: International conference on machine learning, PMLR, 2017, pp. 1263--1272.

\bibitem{khan2024graphmesh}
A.~Khan, M.~Yamada, A.~Chikane, M.~Kaul, Graphmesh: Geometrically generalized mesh refinement using gnns, in: International Conference on Computational Science, Springer, 2024, pp. 120--134.

\bibitem{romit_review}
B.~Sanderse, P.~Stinis, R.~Maulik, S.~E. Ahmed, Scientific machine learning for closure models in multiscale problems: A review, arXiv preprint arXiv:2403.02913 (2024).

\bibitem{kipf2016semi}
T.~N. Kipf, M.~Welling, Semi-supervised classification with graph convolutional networks, arXiv preprint arXiv:1609.02907 (2016).

\bibitem{hamilton2017inductive}
W.~Hamilton, Z.~Ying, J.~Leskovec, Inductive representation learning on large graphs, Advances in neural information processing systems 30 (2017).

\bibitem{gat}
P.~Veli{\v{c}}kovi{\'c}, G.~Cucurull, A.~Casanova, A.~Romero, P.~Lio, Y.~Bengio, Graph attention networks, arXiv preprint arXiv:1710.10903 (2017).

\bibitem{battaglia2018relational}
P.~W. Battaglia, J.~B. Hamrick, V.~Bapst, A.~Sanchez-Gonzalez, V.~Zambaldi, M.~Malinowski, A.~Tacchetti, D.~Raposo, A.~Santoro, R.~Faulkner, et~al., Relational inductive biases, deep learning, and graph networks, arXiv preprint arXiv:1806.01261 (2018).

\bibitem{sanchez2020learning}
A.~Sanchez-Gonzalez, J.~Godwin, T.~Pfaff, R.~Ying, J.~Leskovec, P.~Battaglia, Learning to simulate complex physics with graph networks, in: International conference on machine learning, PMLR, 2020, pp. 8459--8468.

\bibitem{pfaff2020learning}
T.~Pfaff, M.~Fortunato, A.~Sanchez-Gonzalez, P.~Battaglia, Learning mesh-based simulation with graph networks, in: International conference on learning representations, 2020.

\bibitem{fortunato2022multiscale}
M.~Fortunato, T.~Pfaff, P.~Wirnsberger, A.~Pritzel, P.~Battaglia, Multiscale meshgraphnets, arXiv preprint arXiv:2210.00612 (2022).

\bibitem{lino2022multi}
M.~Lino, S.~Fotiadis, A.~A. Bharath, C.~D. Cantwell, Multi-scale rotation-equivariant graph neural networks for unsteady eulerian fluid dynamics, Physics of Fluids 34~(8) (2022).

\bibitem{shivam_gnn_jcp}
S.~Barwey, V.~Shankar, V.~Viswanathan, R.~Maulik, Multiscale graph neural network autoencoders for interpretable scientific machine learning, Journal of Computational Physics 495 (2023) 112537.

\bibitem{deshpande2022magnet}
S.~Deshpande, S.~Bordas, J.~Lengiewicz, Magnet: A graph u-net architecture for mesh-based simulations, arXiv preprint arXiv:2211.00713 (2022).

\bibitem{perera2024multiscale}
R.~Perera, V.~Agrawal, Multiscale graph neural networks with adaptive mesh refinement for accelerating mesh-based simulations, Computer Methods in Applied Mechanics and Engineering 429 (2024) 117152.

\bibitem{lam2023learning}
R.~Lam, A.~Sanchez-Gonzalez, M.~Willson, P.~Wirnsberger, M.~Fortunato, F.~Alet, S.~Ravuri, T.~Ewalds, Z.~Eaton-Rosen, W.~Hu, et~al., Learning skillful medium-range global weather forecasting, Science 382~(6677) (2023) 1416--1421.

\bibitem{li_gnn_md}
Z.~Li, K.~Meidani, P.~Yadav, A.~Barati~Farimani, Graph neural networks accelerated molecular dynamics, The Journal of Chemical Physics 156~(14) (2022).

\bibitem{park2024scalable}
Y.~Park, J.~Kim, S.~Hwang, S.~Han, Scalable parallel algorithm for graph neural network interatomic potentials in molecular dynamics simulations, Journal of chemical theory and computation 20~(11) (2024) 4857--4868.

\bibitem{jaiman_fsi_gnn}
R.~Gao, R.~K. Jaiman, Predicting fluid--structure interaction with graph neural networks, Physics of Fluids 36~(1) (2024).

\bibitem{karthik_neurips_gnn}
J.~Xu, A.~Pradhan, K.~Duraisamy, Conditionally parameterized, discretization-aware neural networks for mesh-based modeling of physical systems, Advances in Neural Information Processing Systems 34 (2021) 1634--1645.

\bibitem{li2024predicting}
T.~Li, S.~Zou, X.~Chang, L.~Zhang, X.~Deng, Predicting unsteady incompressible fluid dynamics with finite volume informed neural network, Physics of Fluids 36~(4) (2024).

\bibitem{jianxun_gnn_galerkin}
H.~Gao, M.~J. Zahr, J.-X. Wang, Physics-informed graph neural galerkin networks: A unified framework for solving pde-governed forward and inverse problems, Computer Methods in Applied Mechanics and Engineering 390 (2022) 114502.

\bibitem{jaiman_phignn}
R.~Gao, I.~K. Deo, R.~K. Jaiman, A finite element-inspired hypergraph neural network: Application to fluid dynamics simulations, Journal of Computational Physics 504 (2024) 112866.

\bibitem{sb_srgnn}
S.~Barwey, P.~Pal, S.~Patel, R.~Balin, B.~Lusch, V.~Vishwanath, R.~Maulik, R.~Balakrishnan, Mesh-based super-resolution of fluid flows with multiscale graph neural networks, Computer Methods in Applied Mechanics and Engineering 443 (2025) 118072.

\bibitem{physgnn}
Y.~Salehi, D.~Giannacopoulos, Physgnn: A physics--driven graph neural network based model for predicting soft tissue deformation in image--guided neurosurgery, Advances in Neural Information Processing Systems 35 (2022) 37282--37296.

\bibitem{schmidt2024towards}
A.~Schmidt, H.~Zunker, A.~Heinlein, M.~J. K{\"u}hn, Towards graph neural network surrogates leveraging mechanistic expert knowledge for pandemic response, arXiv preprint arXiv:2411.06500 (2024).

\bibitem{selvaraju2020grad}
R.~R. Selvaraju, M.~Cogswell, A.~Das, R.~Vedantam, D.~Parikh, D.~Batra, Grad-cam: visual explanations from deep networks via gradient-based localization, International journal of computer vision 128 (2020) 336--359.

\bibitem{ying2019gnnexplainer}
Z.~Ying, D.~Bourgeois, J.~You, M.~Zitnik, J.~Leskovec, Gnnexplainer: Generating explanations for graph neural networks, Advances in neural information processing systems 32 (2019).

\bibitem{luo2020parameterized}
D.~Luo, W.~Cheng, D.~Xu, W.~Yu, B.~Zong, H.~Chen, X.~Zhang, Parameterized explainer for graph neural network, Advances in neural information processing systems 33 (2020) 19620--19631.

\bibitem{huang2022graphlime}
Q.~Huang, M.~Yamada, Y.~Tian, D.~Singh, Y.~Chang, Graphlime: Local interpretable model explanations for graph neural networks, IEEE Transactions on Knowledge and Data Engineering 35~(7) (2022) 6968--6972.

\bibitem{harel2006graph}
J.~Harel, C.~Koch, P.~Perona, Graph-based visual saliency, Advances in neural information processing systems 19 (2006).

\bibitem{zhang2021survey}
Y.~Zhang, P.~Ti{\v{n}}o, A.~Leonardis, K.~Tang, A survey on neural network interpretability, IEEE Transactions on Emerging Topics in Computational Intelligence 5~(5) (2021) 726--742.

\bibitem{barwey2023multiscale}
S.~Barwey, V.~Shankar, V.~Viswanathan, R.~Maulik, Multiscale graph neural network autoencoders for interpretable scientific machine learning, Journal of Computational Physics 495 (2023) 112537.

\bibitem{gao2019graph}
H.~Gao, S.~Ji, Graph u-nets, in: international conference on machine learning, PMLR, 2019, pp. 2083--2092.

\bibitem{barwey2025interpretable}
S.~Barwey, H.~Kim, R.~Maulik, Interpretable a-posteriori error indication for graph neural network surrogate models, Computer Methods in Applied Mechanics and Engineering 433 (2025) 117509.

\bibitem{kucharski2006decadal}
F.~Kucharski, F.~Molteni, A.~Bracco, Decadal interactions between the western tropical pacific and the north atlantic oscillation, Climate dynamics 26 (2006) 79--91.

\bibitem{gao2024predicting}
R.~Gao, R.~K. Jaiman, Predicting fluid--structure interaction with graph neural networks, Physics of Fluids 36~(1) (2024).

\bibitem{velivckovic2019neural}
P.~Veli{\v{c}}kovi{\'c}, R.~Ying, M.~Padovano, R.~Hadsell, C.~Blundell, Neural execution of graph algorithms, arXiv preprint arXiv:1910.10593 (2019).

\bibitem{eliasof2021pde}
M.~Eliasof, E.~Haber, E.~Treister, Pde-gcn: Novel architectures for graph neural networks motivated by partial differential equations, Advances in neural information processing systems 34 (2021) 3836--3849.

\bibitem{yun2019graph}
S.~Yun, M.~Jeong, R.~Kim, J.~Kang, H.~J. Kim, Graph transformer networks, Advances in neural information processing systems 32 (2019).

\bibitem{molteni2003atmospheric}
F.~Molteni, Atmospheric simulations using a gcm with simplified physical parametrizations. i: Model climatology and variability in multi-decadal experiments, Climate Dynamics 20 (2003) 175--191.

\end{thebibliography}

\end{document}